\pdfoutput=1

\documentclass[11pt]{article}

\usepackage[preprint]{acl}

\usepackage{times}
\usepackage{latexsym}

\usepackage[T1]{fontenc}

\usepackage[utf8]{inputenc}

\usepackage{microtype}

\usepackage{inconsolata}

\usepackage[x11names,table]{xcolor}

\usepackage{graphicx}
\definecolor{lightgreen}{rgb}{0.9, 1.0, 0.9}
\newcommand{\colorMbrCbdt}{lightgreen}
\newcommand{\colorOracle}{gray!20}

\usepackage{amsmath}
\usepackage{cleveref}
\usepackage{mathtools}
\usepackage{amsfonts}
\usepackage{bm}
\usepackage{bbm}
\usepackage{booktabs}
\usepackage{tabularx}
\usepackage{nicematrix}
\usepackage{tablefootnote}
\usepackage[subrefformat=parens]{subcaption}
\usepackage{float}

\crefformat{section}{#2Section #1#3}
\crefformat{subsection}{#2Section #1#3}
\crefname{equation}{Equation}{Equation}
\crefname{figure}{Figure}{Figure}
\crefname{table}{Table}{Table}
\crefname{appendix}{Appendix}{Appendix}

\DeclareMathOperator*{\argmax}{argmax}

\DeclareMathOperator*{\expect}{\mathbb{E}}

\newcommand{\stySimilarity}[1]{{#1}}
\newcommand{\similarityMELargeInstruct}{\stySimilarity{mE5${}_\text{large}^\text{instruct}$}}

\newcommand{\similarityME}{\similarityMELargeInstruct{}}
\newcommand{\similarityLaBSE}{\stySimilarity{LaBSE}}
\newcommand{\similarityBM}{\stySimilarity{BM25}}

\newcommand{\styMetric}[1]{\textsc{#1}}
\newcommand{\metricBleu}{\styMetric{Bleu}}
\newcommand{\metricChrf}{\styMetric{chrF}}
\newcommand{\metricComet}{\styMetric{comet}}
\newcommand{\metrickiwi}{\styMetric{kiwi}}
\newcommand{\metricBleurt}{\styMetric{Blrt}}

\newcommand{\metricBertScore}{\styMetric{BS}}

\newcommand{\styDecoding}[1]{{\textsc{#1}}}
\newcommand{\decodingMap}{\styDecoding{Map}}
\newcommand{\decodingMbr}{\styDecoding{Mbr}}
\newcommand{\decodingMcmbr}{\styDecoding{McMbr}}
\newcommand{\decodingQe}{\styDecoding{Qe}}
\newcommand{\decodingCbdt}{\styDecoding{Cbdt}}
\newcommand{\decodingMbrCbdt}{{\substack{\styDecoding{McMbr}\\ \styDecoding{-Cbdt}}}}
\newcommand{\decodingMbrCbdtText}{\styDecoding{Mbr-Cbdt}}
\newcommand{\decodingOracle}{\styDecoding{Oracle}}
\newcommand{\decodingKnn}{$k$\styDecoding{nn-mt}}

\newcommand{\regmark}{${}^{\text{\textregistered}}$}
\newcommand{\trademark}{${}^{\text{\texttrademark}}$}

\newcommand{\styTool}[1]{\texttt{#1}}
\newcommand{\mbrs}{\styTool{mbrs}}
\newcommand{\faiss}{\styTool{Faiss}}
\newcommand{\knntransformers}{\styTool{knn-transformers}}
\newcommand{\bms}{\styTool{BM25S}}

\newcommand{\styVector}[1]{\mathbf{#1}}
\newcommand{\stySet}[1]{\mathcal{#1}}
\newcommand{\styExample}[1]{\tilde{#1}}
\newcommand{\styExampleLoop}[1]{\tilde{#1}'}
\newcommand{\R}{\mathbb{R}}
\newcommand{\N}{\mathbb{N}}
\newcommand{\complexity}[1]{\mathcal{O}\left(#1\right)}
\newcommand{\textInput}{\styVector{x}}
\newcommand{\textOutput}{\styVector{y}}
\newcommand{\textHypothesis}{\styVector{h}}
\newcommand{\textReference}{\styVector{r}}
\newcommand{\modelGenerator}{\theta}
\newcommand{\outputMap}{\styVector{y}^{\decodingMap}}
\newcommand{\outputMbr}{\styVector{y}^{\decodingMbr}}

\newcommand{\outputCbdt}{\styVector{y}^{\decodingCbdt}}
\newcommand{\outputMbrCbdt}{\styVector{y}^{\decodingMbrCbdt}}
\newcommand{\scoreMbr}{U^{\decodingMbr}}
\newcommand{\scoreMcmbr}{U^{\decodingMcmbr}}
\newcommand{\scoreCbdt}{U^{\decodingCbdt}}

\newcommand{\scoreMcmbrNorm}{\bar{U}^{\decodingMcmbr}}
\newcommand{\scoreCbdtNorm}{\bar{U}^{\decodingCbdt}}
\newcommand{\spaceInput}{\stySet{X}}
\newcommand{\spaceOutput}{\stySet{Y}}
\newcommand{\setHypotheses}{\stySet{H}}
\newcommand{\setPseudoRefs}{\hat{\stySet{Y}}}
\newcommand{\vocab}{\stySet{V}}
\newcommand{\setData}{\stySet{D}}
\newcommand{\defFunc}[3]{{#1}\colon{#2}\to{#3}}
\newcommand{\funcUtility}{u}
\newcommand{\setProblems}{\stySet{Q}}
\newcommand{\setActions}{\stySet{A}}
\newcommand{\spaceRewards}{\stySet{R}}
\newcommand{\elmProblem}{q}
\newcommand{\elmAction}{a}
\newcommand{\elmReward}{r}
\newcommand{\setMemory}{\stySet{M}}
\newcommand{\funcSimilarity}{s}
\newcommand{\funcSimilarityNorm}{\bar{s}}

\newcommand{\funcKernelNorm}{\bar{K}}

\newcommand{\numMemHypotheses}{H}
\newcommand{\temperature}{\tau}

\title{
Case-Based Decision-Theoretic Decoding with Quality Memories
}

\author{Hiroyuki Deguchi \and Masaaki Nagata \\
  NTT, Inc. \\
  \texttt{\{hiroyuki.deguchi,masaaki.nagata\}@ntt.com} \\}

\begin{document}
\maketitle
\begin{abstract}

Minimum Bayes risk (MBR) decoding is a decision rule of text generation, which selects the hypothesis that maximizes the expected utility and robustly generates higher-quality texts than maximum a posteriori (MAP) decoding.
However, it depends on sample texts drawn from the text generation model; thus, it is difficult to find a hypothesis that correctly captures the knowledge or information of out-of-domain.
To tackle this issue, we propose case-based decision-theoretic (CBDT) decoding, another method to estimate the expected utility using examples of domain data.
CBDT decoding not only generates higher-quality texts than MAP decoding, but also the combination of MBR and CBDT decoding outperformed MBR decoding in seven domain De--En and Ja$\leftrightarrow$En translation tasks and 
image captioning tasks on MSCOCO and nocaps datasets.
\end{abstract}

\section{Introduction}

Minimum Bayes risk (MBR) decoding robustly generates high-quality texts compared with maximum a posteriori (MAP) decoding, i.e., one-best decoding using beam search~\citep{kumar-byrne-2004-minimum,eikema-aziz-2020-map,eikema-aziz-2022-sampling,freitag-etal-2022-high,fernandes-etal-2022-quality,cheng-vlachos-2023-faster,deguchi-etal-2024-centroid,heineman-etal-2024-improving,lyu-etal-2025-unveiling}.
The key concept of MBR decoding is maximizing the expected utility (EU) of choice from multiple output hypotheses, which is based on EU theory (EUT) in decision-making under uncertainty~\citep{von-neumann-morgenstern-1944-theory}.
EUT is widely applied beyond natural language processing (NLP) to include microeconomics and speech recognition~\citep{conte-etal-2011-mixture,goel-and-byrne-2000-minimum,raina-gales-2023-minimum}.
However, due to a limitation of EUT, if there is a lack of knowledge about the problem or task, it is difficult to accurately estimate the EU, and a hypothesis that reflects preferences cannot be chosen~\citep{gilboa-schmeidler-1995-case}.
Thus, MBR decoding, based on EUT, makes it difficult to generate texts that reflect domain knowledge.

\begin{figure}[t]
    \centering
    \centering
    \includegraphics[width=\linewidth]{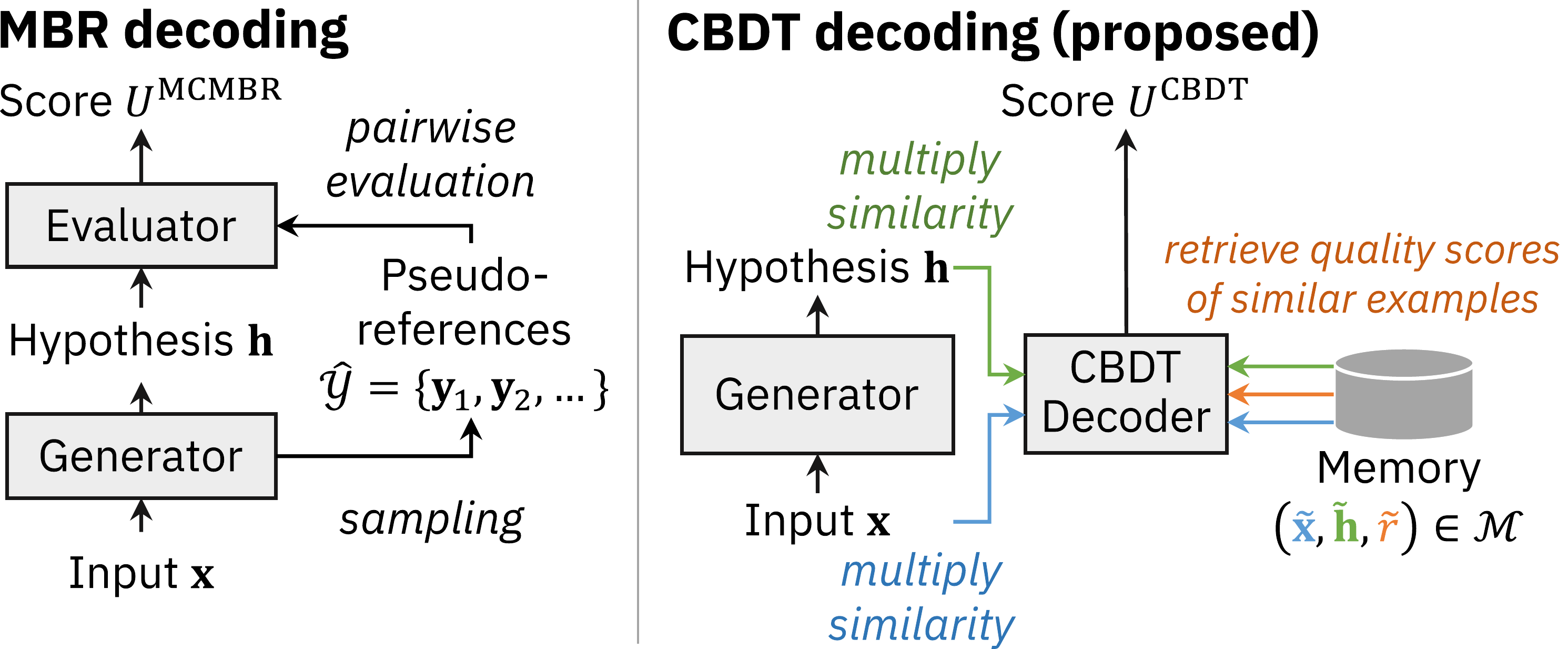}
    \caption{
    MBR decoding (left) and our proposed CBDT decoding (right).
    Both methods select the best hypothesis that maximizes the EU from a hypothesis set.
    ``Generator'' denotes a text generation model, ``Evaluator'' denotes a utility function, i.e., evaluation metric, and ``CBDT Decoder'' computes CBDT scores.
    }
    \label{fig:overview}
\end{figure}

In response to such issues with EUT, in the field of decision theory, case-based decision theory (CBDT) has been proposed, which inductively derives the best action from past experiences~\citep{gilboa-schmeidler-1995-case}.
CBDT evaluates the value of an action on the basis of analogies with similar cases experienced in the past.

To address the limitation of MBR decoding based on EUT, we propose case-based decision-theoretic decoding (CBDT decoding), a novel decision rule for high-quality text generation, which reflects domain-specific preferences by using domain data.
As the preprocessing step, we pre-evaluate the quality scores of multiple generated texts and store them in a ``memory''. 
During generation, CBDT decoding
calculates output scores from the memorized quality scores of similar examples by multiplying the similarity weights between the current problem and memorized examples.
\Cref{fig:overview} shows an overview of MBR decoding and CBDT decoding.
MBR decoding evaluates the quality of a hypothesis using sampled texts called ``pseudo-references'' in decoding time, whereas CBDT decoding retrieves the precomputed quality scores of similar examples.
Notably, CBDT decoding does not depend on ``pseudo-references''; instead, it uses ``true references'' of similar examples.
Moreover, we also propose MBR-CBDT decoding, a combination of MBR decoding and CBDT decoding with score normalization.
Both MBR decoding and CBDT decoding estimate the expected quality but use different information orthogonally, i.e., possibilities and past experiences; thus, combining them further leads to even better quality.

Our CBDT decoding outperformed MAP decoding, and MBR-CBDT decoding generated higher-quality texts compared with MBR decoding in seven domain German--English and Japanese$\leftrightarrow$English translation tasks~\cite{koehn-knowles-2017-six,aharoni-goldberg-2020-unsupervised,nakazawa-etal-2016-aspec,neubig-2011-kftt} and image captioning tasks on MSCOCO~\citep{lin-etal-2014-microsoft,karpathy-and-feifei-2015-deep} and nocaps~\citep{agrawal-etal-2019-nocaps} datasets.

\section{Background}
\paragraph{MBR decoding}
Text generation is a fundamental NLP task that returns the best text given a context.
This paper focuses on decision rules regarding the choice of outputs in conditional text generation.

Let $\textInput \in \spaceInput$ and $\textOutput \in \spaceOutput$ be an input and output text, respectively\footnote{
For simplicity, we formulate an input $\textInput$ as a text, but other modalities such as images can also be used for inputs.
}, where $\spaceInput, \spaceOutput \subseteq \vocab^\ast$ denote the input and output spaces, respectively, and $\vocab^\ast$ denotes the Kleene closure of the vocabulary $\vocab$.
The most widely used text generation method, MAP decoding, finds the most probable text:
\begin{equation}
    \outputMap \coloneqq \argmax_{\textHypothesis \in \spaceOutput} p\left(\textHypothesis | \textInput; \modelGenerator\right),
\end{equation}
where $\modelGenerator$ is a text generation model.
Because $\spaceOutput$ is an infinite set, the output is selected from a set of hypotheses $\setHypotheses \subset \spaceOutput$ instead of $\spaceOutput$.

As a more quality-aware decision strategy\footnote{
MAP decoding and MBR decoding are equivalent when {$u$} is an indicator function {$\mathbbm{1}_{\textHypothesis = \textOutput}$}; thus, MAP decoding can be regarded as a special case of MBR decoding.
}, MBR decoding selects the hypothesis $\outputMbr$ that maximizes the expected utility (EU):
\begin{align}
    \outputMbr &\coloneqq \argmax_{\textHypothesis \in \setHypotheses} \scoreMbr(\textHypothesis; \textInput)  \\
    &= \argmax_{\textHypothesis \in \setHypotheses} \expect_{\textOutput \sim \Pr(\cdot|\textInput)} \left[ \funcUtility(\textHypothesis, \textOutput) \right].
\end{align}
A reference text $\textOutput \in \spaceOutput$  occurs according to the true output distribution $\Pr(\cdot|\textInput)$, and $\defFunc{u}{\spaceOutput \times \spaceOutput}{\R}$ denotes the utility function that satisfies $\textHypothesis \succeq \textHypothesis' \iff \funcUtility(\textHypothesis, \textReference) \geq \funcUtility(\textHypothesis', \textReference)$,  where $\succeq$ denotes the preference relation.
For a utility function, an evaluation metric of output quality is often employed.
Here, $\Pr(\cdot|\textInput)$ is unknown; thus, the EU $\scoreMbr$ is typically estimated using the Monte Carlo (MC) method~\citep{eikema-aziz-2020-map,eikema-aziz-2022-sampling}:
\begin{align}
    \scoreMcmbr(\textHypothesis; \setPseudoRefs) &\coloneqq \frac{1}{|\setPseudoRefs|}\sum_{\textOutput \in \setPseudoRefs} \funcUtility(\textHypothesis, \textOutput),
\end{align}
where $\setPseudoRefs \coloneqq \{ \textOutput_i \}_{i=1}^{|\setPseudoRefs|} \sim p(\textOutput|\textInput; \modelGenerator)$ are pseudo-references, a multiset (a.k.a. bag) of sampled texts that are drawn from the text generation model $\modelGenerator$.

The decision of MBR decoding highly depends on the distribution of pseudo-references~\citep{ohashi-etal-2024-true,kamigaito-etal-2025-diversity}.
Hence, in domains where the text generation model lacks knowledge, the EU estimation could be unreliable due to the discrepancy in distribution between the pseudo-references and true references, making it difficult to select the hypothesis that reflects domain-specific knowledge and information.

\paragraph{Case-based decision theory}
In decision theory, case-based decision theory (CBDT), which derives the best action from past experiences, has been proposed~\citep{gilboa-schmeidler-1995-case}.
Decision-makers following CBDT choose actions on the basis of the rewards of similar experienced examples.
In CBDT, the set of examples is defined by the triplet: a set of problems $\setProblems$, a set of actions $\setActions$, and the reward space $\spaceRewards$.
Let $\elmProblem \in \setProblems$ denote the problem currently being faced.
Decision-makers following CBDT choose an action $\elmAction^\star \in \setActions$ on the basis of the memory $\setMemory \subseteq \setProblems \times \setActions \times \spaceRewards$, a set of examples they have experienced in the past, as follows:
\begin{equation}
    \elmAction^\star \coloneqq \argmax_{\elmAction \in \setActions} \sum_{(\styExample{\elmProblem}, \styExample{\elmAction}, \styExample{\elmReward}) \in \setMemory} \funcSimilarity(\elmProblem, \styExample{\elmProblem}) \mathbbm{1}_{\elmAction=\styExample{\elmAction}} \styExample{\elmReward},
    \label{eq:CBDT:score}
\end{equation}
where $\defFunc{\funcSimilarity}{\setProblems \times \setProblems}{[0, 1]}$ is the similarity between problems.
From \cref{eq:CBDT:score}, CBDT selects the action that maximizes the sum of the rewards weighted by the similarity between the current facing problem and experienced problems.

\section{Proposed Method: CBDT Decoding}
\begin{figure}[t]
    \centering
    \includegraphics[width=\linewidth]{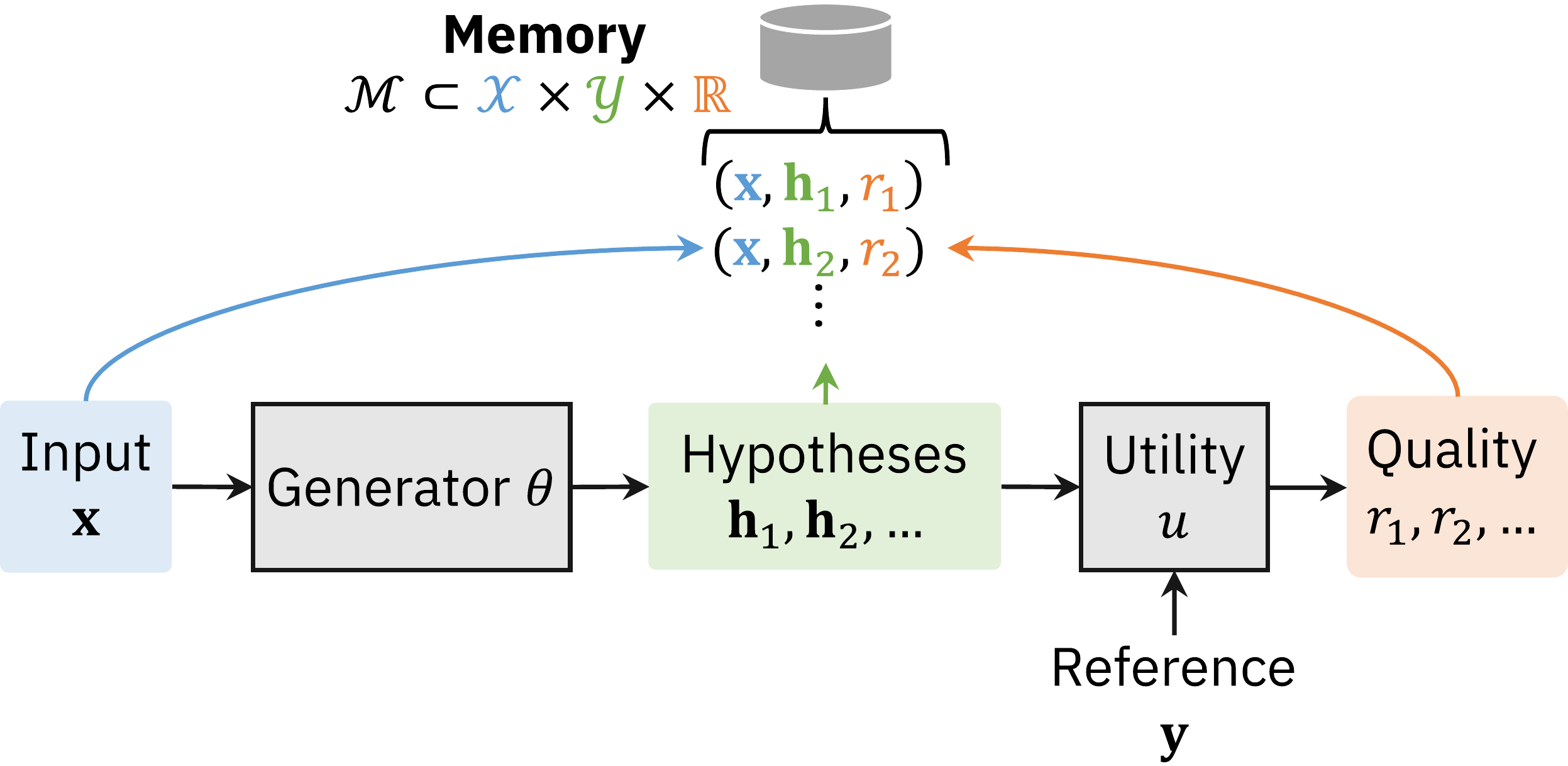}
    \caption{Memorization of CBDT decoding.}
    \label{fig:proposal:memory}
\end{figure}

\begin{figure}[t]
    \centering
    \includegraphics[width=1.0\linewidth]{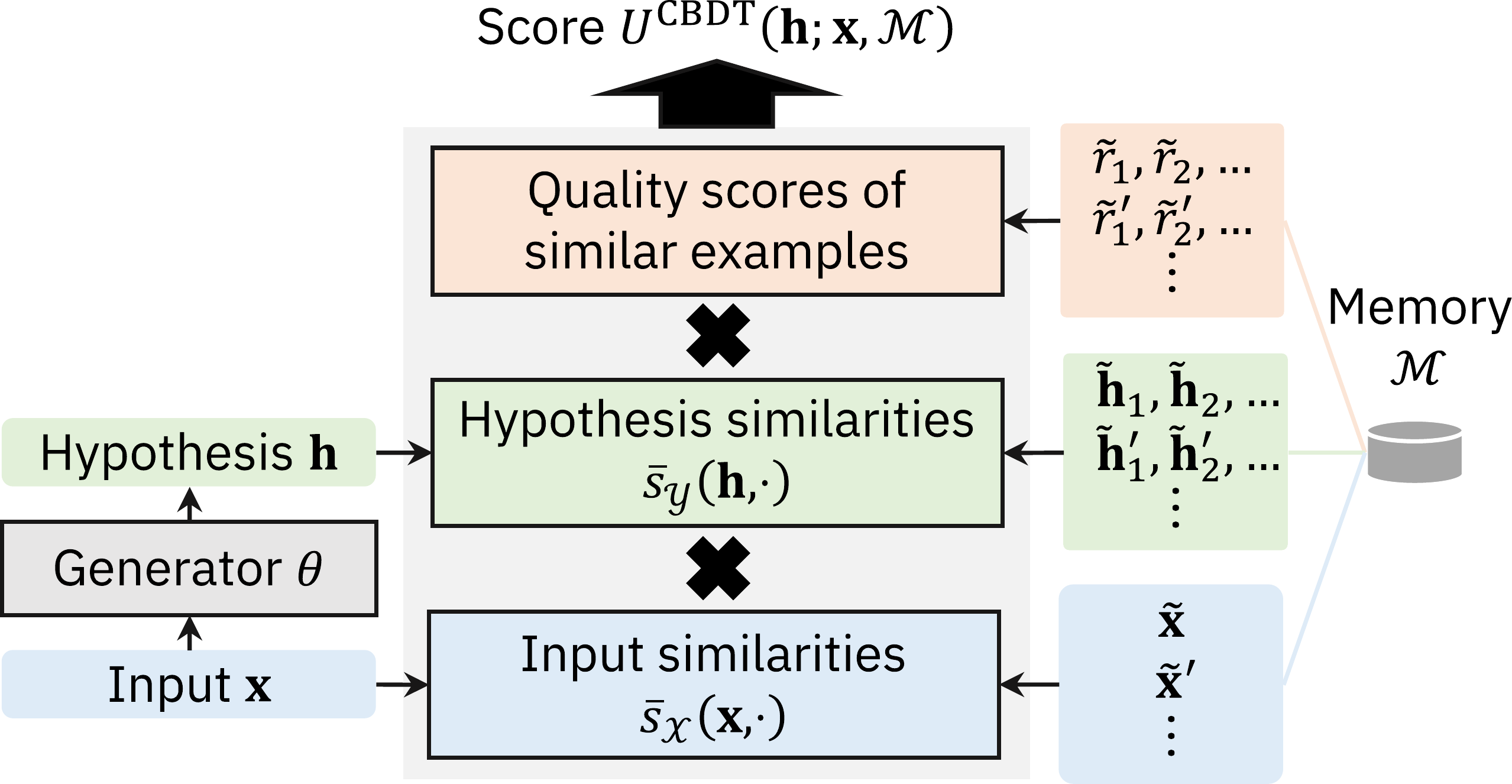}
    \caption{
    CBDT score calculation for a hypothesis $\textHypothesis \in \setHypotheses$.
    CBDT decoding selects the hypothesis that maximizes the score $\scoreCbdt$ from hypotheses $\setHypotheses$.
    }
    \label{fig:proposal:score}
\end{figure}

We propose \emph{case-based decision-theoretic (CBDT) decoding} for high-quality text generation utilizing domain data.
It pre-evaluates and stores the rewards of hypothesis selection (\cref{fig:proposal:memory}) and decides the output referring to the memorized information (\cref{fig:proposal:score}).
For the purpose of text generation, we hereafter redefine the problem set $\setProblems$ as the input space $\spaceInput$, action set $\setActions$ as the output space $\spaceOutput$, and reward space $\spaceRewards$ as the quality scores of output texts calculated with the utility function $\funcUtility$.

\subsection{CBDT decoding}
\paragraph{Memorization}

We first construct a \emph{memory} from parallel data consisting of pairs of input and its reference output texts $\setData \coloneqq \{ (\textInput_i, \textOutput_i) \}_{i=1}^{|\setData|}$.
We generate sets of hypotheses $\setHypotheses_{\textInput} \subset \spaceOutput$ with $\numMemHypotheses \in \N$ hypotheses for each input $\textInput$ in the parallel data $\setData$.
\begin{equation}
    \setHypotheses_{\textInput} \coloneqq \left\{ \textHypothesis_\ell \right\}_{\ell=1}^\numMemHypotheses \sim p(\cdot|\textInput; \modelGenerator).
\end{equation}
Since $\setHypotheses_{\textInput}$ is a set, $|\setHypotheses_{\textInput}| \leq H$, i.e., the elements are deduplicated.
Then, we evaluate each generated hypothesis using the reference text $\textOutput$ and store the triplets in the memory $\setMemory \subseteq \spaceInput \times \spaceOutput \times \R$ as follows:
\begin{align}
    \setMemory \coloneqq &\{ \left(\textInput, \textHypothesis_i, \elmReward_i
 \right) \mid \textHypothesis_i \in \setHypotheses_{\textInput}, (\textInput, \textOutput) \in \setData \}, \\
 &\text{where~} \elmReward_i = \funcUtility(\textHypothesis_i, \textOutput).
\end{align}

\paragraph{Decoding}
CBDT decoding decides the output referring to the preconstructed memory.
The na\"ive CBDT decoding 
based on \Cref{eq:CBDT:score} selects the hypothesis that maximizes the following score:
\begin{equation}
    \label{eq:CBDT:naive}
    U^{\substack{\styDecoding{Cbdt}\\ \styDecoding{Na\"ive}}}(\textHypothesis; \textInput, \setMemory) \coloneqq \sum_{(\styExample{\textInput}, \styExample{\textHypothesis}, \styExample{\elmReward}) \in \setMemory} \funcSimilarity(\textInput, \styExample{\textInput}) \mathbbm{1}_{\textHypothesis = \styExample{\textHypothesis}} \styExample{\elmReward}.
\end{equation}
There are two problems with na\"ive CBDT decoding.
One is that the similarity function $\funcSimilarity$ must return values within the range $[0, 1]$; thus, arbitrary similarity functions cannot be used.
The other is due to the sparsity of natural language data.
If the hypothesis $\textHypothesis$ is not contained in $\setMemory$, i.e., if the model does not generate exactly the same text as $\textHypothesis$ in memorization, $U^{\substack{\styDecoding{Cbdt}\\ \styDecoding{Na\"ive}}}$ always returns $0$ because of the indicator function $\mathbbm{1}_{\textHypothesis=\styExample{\textHypothesis}}$.
To solve these problems, we instead use the normalized similarity, and also introduce the similarity between hypotheses $\textHypothesis$ and $\styExample{\textHypothesis}$ instead of the indicator function to soften the equivalence checking.
\begin{align}
    &\scoreCbdt(\textHypothesis; \textInput, \setMemory) \nonumber \\
    &\coloneqq \sum_{(\styExample{\textInput}, \styExample{\textHypothesis}, \styExample{\elmReward}) \in \setMemory} \funcSimilarityNorm_\spaceInput(\textInput, \styExample{\textInput}; \setMemory) \funcSimilarityNorm_\spaceOutput(\textHypothesis, \styExample{\textHypothesis}; \setHypotheses_{\styExample{\textInput}}) \styExample{\elmReward}.
\end{align}
$\defFunc{\funcSimilarityNorm_\spaceInput}{\spaceInput \times \spaceInput}{[0, 1]}$
and $\defFunc{\funcSimilarityNorm_\spaceOutput}{\spaceOutput \times \spaceOutput}{[0, 1]}$ 
are the normalized similarities for input and output sides, respectively, and we formulate them as:
\begin{align}
    &\funcSimilarityNorm_\spaceInput(\textInput, \styExample{\textInput}; \setMemory) \coloneqq 
    \frac{
        \exp{
            \frac{
                \funcSimilarity_\spaceInput(\textInput, \styExample{\textInput})
            }{
                \temperature_\spaceInput
            }
        }
    }{
        {
            \displaystyle
            \sum_{(\styExampleLoop{\textInput}, \styExampleLoop{\textHypothesis}, \styExampleLoop{\elmReward}) \in \setMemory}
        }
        \exp{
            \frac{
                \funcSimilarity_\spaceInput(\textInput, \styExampleLoop{\textInput})
            }{
                \temperature_\spaceInput
            }
        }
    }, \\
    &\funcSimilarityNorm_\spaceOutput(\textHypothesis, \styExample{\textHypothesis}; \setHypotheses_{\styExample{\textInput}}) \coloneqq 
    \frac{
        \exp{
            \frac{
                \funcSimilarity_\spaceOutput(\textHypothesis, \styExample{\textHypothesis})
            }{
                \temperature_\spaceOutput
            }
        }
    }{
        \sum_{\styExampleLoop{\textHypothesis} \in \setHypotheses_{\styExample{\textInput}}}
        \exp{
            \frac{
                \funcSimilarity_\spaceOutput(\textHypothesis, \styExampleLoop{\textHypothesis})
            }{
                \temperature_\spaceOutput
            }
        }
    },
\end{align}
where $\defFunc{\funcSimilarity_\spaceInput}{\spaceInput \times \spaceInput}{\R}$
and $\defFunc{\funcSimilarity_\spaceOutput}{\spaceOutput \times \spaceOutput}{\R}$
are arbitrary similarity functions for the input space and output space, respectively, and $\tau_\spaceInput$ and $\tau_\spaceOutput$ are the temperatures for the similarities.

Our proposed \emph{CBDT decoding} first retrieves the $k$-nearest neighbor examples $\hat{\setMemory} \subseteq \setMemory$ based on the input side similarity $\funcSimilarity_\spaceInput(\textInput, \styExample{\textInput})$ to reduce the space complexity\footnote{
This is because the similarity matrix between the current hypotheses $\setHypotheses$ and memorized hypotheses $\{\styExample{\textHypothesis} \mid (\styExample{\textInput}, \styExample{\textHypothesis}, \styExample{\elmReward}) \in \setMemory \}$ is often space-consuming.
If all the memorized examples are used, the space complexity will be $\complexity{|\setHypotheses||\setMemory|} = \complexity{|\setHypotheses||\setData|\numMemHypotheses}$.
For instance, if $|\setHypotheses| = 1,024$, $|\setData| = 100,000$, and $\numMemHypotheses=256$, the size of the similarity matrix will be $1,024 \times 100,000 \times 256 \times 32~\text{bit} \simeq 97.7~\text{GiB}$.
}, then selects the hypothesis that maximizes the score $\scoreCbdt$:
\begin{equation}
    \label{eq:CBDT}
    \outputCbdt \coloneqq \argmax_{\textHypothesis \in \setHypotheses} \scoreCbdt(\textHypothesis; \textInput, \hat{\setMemory}).
\end{equation}
Note that $|\hat{\setMemory}| \leq \numMemHypotheses k$ since the memory has at most $\numMemHypotheses$ hypotheses $\setHypotheses_{\styExample{\textInput}}$ for each input text $\styExample{\textInput}$.

\subsection{Interpolation of MBR decoding and CBDT decoding}

Both MBR and CBDT decoding aim to find the hypothesis that maximizes the utility, but they use orthogonal approaches to estimate the EU.
Intuitively, MBR decoding mainly focuses on possibilities
and CBDT decoding utilizes past experiences.
Now, we propose \emph{MBR-CBDT decoding}, which combines them, to further improve the quality:
\begin{align}
    \outputMbrCbdt \coloneqq \argmax_{\textHypothesis \in \setHypotheses} {} &(1-\lambda) \scoreMcmbrNorm(\textHypothesis; \setPseudoRefs) \nonumber \\
    &+ \lambda\scoreCbdtNorm(\textHypothesis; \textInput, \hat{\setMemory}), \label{eq:mbr-cbdt}
\end{align}
where $\lambda \in [0, 1]$ balances two terms.
To adjust the range of scores, $\scoreMcmbrNorm$ and $\scoreCbdtNorm$ normalize the scores of $\scoreMcmbr$ and $\scoreCbdt$, respectively, by min-max normalization over the hypothesis set $\setHypotheses$.

\subsection{Fast similarity calculation}
\label{sec:prop:fastsim}
CBDT decoding calculates similarities between the current input/hypothesis and memorized inputs/hypotheses during decoding, respectively, which consumes time linearly proportional to the number of examples.
Fortunately, many similarity functions, including BM25~\citep{jones-etal-2000-probabilistic} and contextualized text embeddings, can prevent slowdown in decoding speed by precomputing the statistical information or embeddings on the example side in advance.
Note that CBDT decoding only uses neighboring examples as described in \cref{eq:CBDT}; thus, it does not need to load the cache of all examples into the RAM.

\section{Experiments}
\label{sec:exp}
We evaluated our methods in translation and image captioning tasks.
To compare the pure decoding performance, we evaluated the output quality of each decoding method without additional model training.
Specifically, we compared MAP decoding (\decodingMap), N-best reranking using a quality estimation model (\decodingQe),
MBR decoding with MC estimation (\decodingMbr), CBDT decoding (\decodingCbdt), and MBR-CBDT decoding (\decodingMbrCbdtText) with $\lambda=0.5$.
We also compared them with oracle (\decodingOracle{}), which selects the hypotheses using the reference texts.
For the translation tasks, we evaluated $k$-nearest neighbor machine translation (\decodingKnn{})~\citep{khandelwal-etal-2021-nearest}, a baseline of example-based decoding.
We implemented our methods using \mbrs{}~\citep{deguchi-etal-2024-mbrs}, and used \knntransformers~\citep{alon-etal-2022-neuro} for \decodingKnn{}.

\subsection{Machine translation}
\label{sec:exp:translation}

\begin{table*}[t]
    \centering
    \tabcolsep 2pt
    \begin{tabular}{@{}c@{}}
    \begin{minipage}{\linewidth}
    \centering
    \small
    \begin{NiceTabular}{@{}lrrr rrr rrr rrr rrr@{}}[colortbl-like]
    \toprule
        & \multicolumn{3}{c}{IT} & \multicolumn{3}{c}{Koran} &\multicolumn{3}{c}{Law} &  \multicolumn{3}{c}{Medical} & \multicolumn{3}{c}{Subtitles}\\
        \cmidrule(lr){2-4}\cmidrule(lr){5-7}\cmidrule(lr){8-10}\cmidrule(lr){11-13}\cmidrule(l){14-16}
        Decoding & \metricChrf & \metricComet & \metricBleurt{} & \metricChrf & \metricComet & \metricBleurt{} & \metricChrf & \metricComet & \metricBleurt{} & \metricChrf & \metricComet & \metricBleurt{} & \metricChrf & \metricComet & \metricBleurt{} \\
        \midrule
        \decodingMap & 45.5 & 76.1 & 58.3 & 23.7 & 57.9 & 46.1 & 48.5 & 74.6 & 61.8 & 50.7 & 78.0 & 61.8 & 39.8 & 73.5 & 55.5 \\
        \decodingQe & 51.0 & 79.1 & 58.9 & 36.2 & 71.9 & \textbf{50.4} & 58.3 & 84.0 & 66.4 & 55.0 & 81.6 & 62.5 & 40.6 & 77.1 & \underline{56.4} \\
        \decodingKnn & 50.2 & 79.8 & \textbf{61.4} & 28.0 & 63.7 & 46.2 & 58.6 & 82.5 & \underline{66.7} & \underline{57.3} & 81.7 & \textbf{64.9} & 42.1 & 74.7 & 56.1 \\
        \midrule
        \multicolumn{16}{@{}l@{}}{Utility $\funcUtility$: \metricChrf{}} \\
        \decodingMbr{} & 52.7 & 77.8 & 58.7 & \underline{37.3} & 68.1 & 49.6 & \underline{60.1} & 82.3 & 65.8 & 56.7 & 80.3 & 62.6 & \underline{42.4} & 74.4 & 56.1 \\
        \decodingCbdt{} & 51.7 & 77.1 & 57.7 & 34.0 & 63.0 & 46.3 & 58.9 & 80.4 & 63.9 & 56.2 & 78.4 & 60.1 & 39.3 & 71.7 & 53.0 \\
        \rowcolor{lightgreen}
        \decodingMbrCbdtText{} & \textbf{54.6} & 78.7 & \underline{59.8} & \textbf{37.5} & 67.4 & 49.2 & \textbf{61.6} & 82.4 & 66.2 & \textbf{58.4} & 80.4 & 62.8 & \textbf{43.3} & 74.5 & 55.8 \\
        \rowcolor{\colorOracle}
        \decodingOracle{} & 63.9 & 82.4 & 66.9 & 48.0 & 69.8 & 53.3 & 69.5 & 84.1 & 70.0 & 67.0 & 82.4 & 67.5 & 57.1 & 77.3 & 61.8 \\
        \midrule
        \multicolumn{16}{@{}l@{}}{Utility $\funcUtility$: \metricComet} \\
        \decodingMbr{} & 51.6 & \underline{81.1} & 59.7 & 35.9 & \textbf{73.5} & \underline{50.2} & 58.4 & \underline{84.6} & 66.4 & 56.0 & \underline{82.9} & 63.5 & 41.6 & \underline{77.9} & \textbf{56.6} \\
        \decodingCbdt{} & 50.6 & 79.7 & 59.3 & 33.6 & 68.1 & 48.0 & 57.0 & 82.7 & 64.7 & 54.0 & 81.1 & 60.9 & 38.3 & 74.9 & 54.0 \\
        \rowcolor{lightgreen}
        \decodingMbrCbdtText{} & \underline{52.9} & \textbf{82.2} & \textbf{61.4} & 35.9 & \underline{73.3} & \textbf{50.4} & 59.2 & \textbf{85.0} & \textbf{67.2} & 56.5 & \textbf{83.3} & \underline{63.6} & 41.1 & \textbf{78.2} & \underline{56.4} \\
        \rowcolor{\colorOracle}
        \decodingOracle{} & 58.5 & 86.5 & 67.6 & 39.8 & 77.8 & 53.9 & 63.7 & 87.2 & 70.7 & 61.6 & 86.0 & 68.3 & 50.3 & 83.1 & 62.3 \\
        \bottomrule
    \end{NiceTabular}
    \subcaption{The translation quality in the five domain De--En translation tasks.}
    \end{minipage} \\
    \\
    \begin{minipage}{\linewidth}
    \tabcolsep 3.5pt
    \centering
    \small
    \begin{NiceTabular}{@{}lrrr rrr rrr rrr@{}}[colortbl-like]
    \toprule
        & \multicolumn{6}{c}{ASPEC} & \multicolumn{6}{c}{KFTT} \\
        \cmidrule(lr){2-7}\cmidrule(l){8-13}
        & \multicolumn{3}{c}{Ja--En} & \multicolumn{3}{c}{En--Ja} &\multicolumn{3}{c}{Ja--En} &  \multicolumn{3}{c}{En--Ja} \\
        \cmidrule(lr){2-4}\cmidrule(lr){5-7}\cmidrule(lr){8-10}\cmidrule(l){11-13}
        Decoding & \metricChrf{} & \metricComet{} & \metricBleurt{} & \metricChrf{} & \metricComet{} & \metricBleurt{} & \metricChrf{} & \metricComet{} & \metricBleurt{} & \metricChrf{} & \metricComet{} & \metricBleurt{}  \\
        \midrule
        \decodingMap{} & 34.9 & 65.8 & 49.5 & 14.0 & 69.6 & 44.9 & 13.6 & 40.0 & 31.6 & 8.4 & 57.7 & 31.9 \\
        \decodingQe{} & 45.0 & 79.4 & \underline{56.2} & 19.8 & 85.1 & \underline{53.7} & 28.4 & 63.8 & \textbf{40.1} & 13.8 & 73.0 & 37.8 \\
        \decodingKnn{} & 42.8 & 75.0 & 53.2 & 19.9 & 81.8 & 52.4 & 22.0 & 56.1 & 36.3 & 12.5 & 69.5 & 37.0 \\
        \midrule
        \multicolumn{13}{@{}l@{}}{Utility $\funcUtility$: \metricChrf} \\
        \decodingMbr{} & \underline{47.2} & 76.5 & 54.2 & 19.7 & 81.8 & 51.8 & \underline{30.7} & 57.6 & 37.1 & 13.8 & 70.1 & 37.0 \\
        \decodingCbdt{} & 46.3 & 76.4 & 53.9 & \underline{20.2} & 83.0 & 52.9 & 30.3 & 59.3 & 37.4 & \underline{14.0} & 70.0 & 36.9 \\
        \rowcolor{\colorMbrCbdt}
        \decodingMbrCbdtText{} & \textbf{48.3} & 77.6 & 55.4 & \textbf{20.8} & 83.4 & \underline{53.7} & \textbf{31.9} & 59.1 & 37.9 & \textbf{14.5} & 71.2 & 38.0 \\
        \rowcolor{\colorOracle}
        \decodingOracle{} & 57.4 & 78.0 & 57.1 & 31.0 & 83.3 & 55.7 & 39.2 & 60.9 & 40.3 & 21.2 & 71.5 & 40.3 \\
        \midrule
        \multicolumn{13}{@{}l@{}}{Utility $\funcUtility$: \metricComet} \\
        \decodingMbr{} & 45.2 & \underline{80.2} & 55.4 & 19.5 & \underline{86.4} & 53.4 & 28.6 & \underline{64.3} & 38.3 & 13.8 & \underline{75.5} & 37.8 \\
        \decodingCbdt{} & 44.6 & 78.2 & 55.0 & 19.1 & 84.2 & 53.4 & 28.3 & 63.2 & 38.3 & 13.6 & 73.0 & \underline{38.5} \\
        \rowcolor{\colorMbrCbdt}
        \decodingMbrCbdtText{} & 45.8 & \textbf{80.4} & \textbf{56.3} & 20.1 & \textbf{86.5} & \textbf{54.7} & 29.2 & \textbf{66.0} & \underline{39.4} & \underline{14.0} & \textbf{76.0} & \textbf{39.2} \\
        \rowcolor{\colorOracle}
        \decodingOracle{} & 49.5 & 82.8 & 58.6 & 23.2 & 88.7 & 57.1 & 30.9 & 70.7 & 41.6 & 15.2 & 80.2 & 41.0 \\
        \bottomrule
    \end{NiceTabular}
    \subcaption{The translation quality in the two domain Ja$\leftrightarrow$En translation tasks.}
    \end{minipage}
    \end{tabular}
    \caption{
    Translation quality in the seven domain translation tasks.
    \colorbox{lightgreen}{Green rows} show the results of \decodingMbrCbdtText{}, and \colorbox{\colorOracle}{gray rows} show the results of \decodingOracle{}.
    The \textbf{bold} and \underline{underlined} scores indicate the best and second-best scores in each column except for \decodingOracle{}, respectively.
    All scores are shown as percentages (\%).
    }
    \label{tab:results:domain_translation}
\end{table*}

\begin{table}[t]
    \centering
    \small
    \begin{tabular}{@{}lrrrr@{}}
        \toprule
        Decoding & Avg & SD & Min & Max \\
        \midrule
        \decodingQe{} & 363.5 & $\pm$ 0.4 & 362.9 & 364.0 \\
        \decodingKnn{} & 2972.9 & $\pm$ 5.8 & 2965.4 & 2981.3 \\
        \midrule
        Utility $\funcUtility$: \metricChrf{} \\
        \decodingMbr{} & 6400.3 & $\pm$ 104.7 & 6192.1 & 6463.8 \\
        \decodingCbdt{} & 120.2 & $\pm$ 0.8 & 119.0 & 121.1 \\
        \decodingMbrCbdtText{} & 6521.6 & $\pm$ 105.3 & 6312.2 & 6585.2 \\
        \midrule
        Utility $\funcUtility$: \metricComet{} \\
        \decodingMbr{} & 899.8 & $\pm$ 1.2 & 898.8 & 902.0 \\
        \decodingCbdt{} & 158.6 & $\pm$ 3.8 & 155.6 & 165.9 \\
        \decodingMbrCbdtText{} & 1087.1 & $\pm$ 3.8 & 1084.2 & 1094.5 \\
        \bottomrule
    \end{tabular}
    \caption{
    Running times per test case (msec) over 5 runs.
    Columns ``Avg'', ``SD'', ``Min'', and ``Max'' indicate average, standard deviation, minimum, and maximum running times, respectively.
    $\decodingCbdt{}$ and $\decodingMbrCbdtText{}$ include times of encoding texts and calculating similarity.
    }
    \label{tab:results:speed:it}
\end{table}

\paragraph{Setup}
We conducted German--English (De--En) translation experiments in five domains: IT, Koran, law, medical, and subtitles~\citep{koehn-knowles-2017-six,aharoni-goldberg-2020-unsupervised}, and Japanese$\leftrightarrow$English (Ja$\leftrightarrow$En) translation experiments in two domains: scientific paper (ASPEC)~\citep{nakazawa-etal-2016-aspec} and Wikipedia's Kyoto articles (KFTT)~\citep{neubig-2011-kftt}.
We generated 1,024 translation hypotheses for each input using M2M100\footnote{
\texttt{facebook/m2m100\_418M}
}~\citep{fan-etal-2021-beyond} via epsilon sampling with $\varepsilon=0.02$~\citep{freitag-etal-2023-epsilon}.
We used the hypothesis set for the pseudo-references.
In \decodingKnn{}, we stored the input representations of the final feed-forward layer in the decoder, retrieved the top-64 nearest neighbor tokens using \faiss{}~\citep{johnson-etal-2019-billion,douze-etal-2024-faiss}, and interpolated the output probability with the temperature of 100.0 and $\lambda=0.5$~\citep{khandelwal-etal-2021-nearest}.
For \decodingCbdt{} and \decodingMbrCbdtText{}, the memories were constructed from parallel data of each domain, and we generated $\numMemHypotheses=256$ hypotheses for each example in De--En and $\numMemHypotheses=64$ in Ja$\leftrightarrow$En.
In ASPEC, we only used the top 1 million translation pairs from the training set to avoid noisy data, following~\citet{nakazawa-etal-2016-aspec}.
For the similarity functions of input and output texts, we used the cosine similarity of sentence embeddings of \texttt{intfloat/multingual-e5-large-instruct}\footnote{
We used the instruction for the semantic textual similarity (STS) task as follows:
``\texttt{Instruct: Retrieve semantically similar text.\textbackslash{}nQuery: }''
} (\similarityME{}) \citep{wang-etal-2024-multilingual}.
$\scoreCbdt$ was calculated using examples that have the $k=256$ nearest neighbor input texts, i.e., using $\numMemHypotheses k = 65{,}536$ triplets at most in De--En.
We set the temperature parameters to $\temperature_\spaceInput=0.01$ and $\temperature_\spaceOutput=0.01$ for De--En, and $\temperature_\spaceInput=0.1$ and $\temperature_\spaceOutput=0.01$ for Ja$\leftrightarrow$En, respectively\footnote{
The details of tuning are described in \cref{sec:tuning}.
}.
We used \metricChrf{} \citep{popovic-2015-chrf} and \metricComet{}\footnote{
\texttt{Unbabel/wmt22-comet-da}
}~\cite{rei-etal-2022-comet} for the utility function, and CometKiwi\footnote{
\texttt{Unbabel/wmt22-cometkiwi-da}
} (\metrickiwi{}) \citep{rei-etal-2022-cometkiwi} for the QE model.
We evaluated the translation quality on \metricChrf{}, \metricComet{}, and BLEURT (\metricBleurt{}) \citep{sellam-etal-2020-bleurt}.
We also measured the execution time on a 32-core Intel\regmark{} Xeon\regmark{} Gold 6426Y and a single NVIDIA RTX\trademark{} 6000 Ada.
We calculated utility scores and similarity with a batch size of 256 sentences.

\paragraph{Translation quality}
\Cref{tab:results:domain_translation} demonstrates the results of the seven domain translation tasks in De--En and Ja$\leftrightarrow$En.
In all translations, \decodingCbdt{} improved up to 16.7\%
 +11.3\% in \metricChrf{} and +23.2\% in \metricComet{} compared with \decodingMap{}.
In addition, \decodingMbrCbdtText{} outperformed \decodingMbr{} in the given utility.
Specifically, it improved up to +1.9\% in \metricChrf{} and +1.7\% in \metricComet{}.
Moreover, it also achieved the best \metricBleurt{} scores in six of the nine test sets, even though it was a non-target utility.

\paragraph{Decoding speed}
We measured the running time of hypothesis selection in the IT domain.
\Cref{tab:results:speed:it} shows the statistics of running time per test case over 5 runs.
\decodingCbdt{} took less than 0.2 second regardless of utilities, whereas \decodingMbr{} took 0.9 second when using \metricComet{} and more than 6.4 seconds when using \metricChrf{}.
CBDT decoding is designed so that the decoding speed does not depend on the cost of the utility function.
This is because \decodingMbr{} evaluates each hypothesis using multiple pseudo-references, i.e., it requires quadratic time, whereas \decodingCbdt{} does not call the utility function during decoding.
One of the bottlenecks of \decodingCbdt{} is text encoding, but by leveraging the proposed method described in \cref{sec:prop:fastsim}, only the query, consisting of a single input and $|\setHypotheses|$ hypotheses, needs to be encoded during decoding.
Thus, the computational cost remains relatively low.
The running time of \decodingMbrCbdtText{} is the sum of that of \decodingMbr{} and a small utility-independent overhead incurred by \decodingCbdt{}.

\subsection{Image captioning}
\label{sec:exp:caption}
To evaluate the effectiveness of our methods in multimodal tasks, we experimented in the image captioning task.
In this task, the input is an image instead of text.
Note that our methods do not require cross-modal embeddings for similarity calculation because the similarities of the input and output sides are calculated independently.

\begin{table}[t]
    \centering
    \small
    \tabcolsep 4pt
    \begin{NiceTabular}{@{}l rrr rrr@{}}[colortbl-like]
        \toprule
         & \multicolumn{3}{c}{MSCOCO} & \multicolumn{3}{c}{nocaps} \\
         \cmidrule(lr){2-4} \cmidrule(l){5-7}
         Decoding & \metricBleu{} & \metricChrf{} & \metricBertScore{} & \metricBleu{} & \metricChrf{} & \metricBertScore{} \\
         \midrule
         \decodingMap{} & 17.5 & 35.1 & 64.5 & 6.2 & 20.4 & 53.6 \\
        \decodingMbr{} & \underline{25.6} & \underline{43.2} & \underline{68.4} & \underline{26.3} & \underline{40.6} & \underline{65.8} \\
        \decodingCbdt{} & 19.5 & 41.4 & 66.3 & 20.1 & 37.3 & 63.5 \\
        \rowcolor{\colorMbrCbdt}
        \decodingMbrCbdtText{} & \textbf{26.0} & \textbf{44.4} & \textbf{68.6} & \textbf{26.8} & \textbf{41.6} & \textbf{66.3} \\
        \rowcolor{\colorOracle}
        \decodingOracle{} & 39.4 & 53.2 & 72.4 & 39.2 & 50.8 & 70.3 \\
         \bottomrule
    \end{NiceTabular}
    \caption{
    Results of image captioning tasks on the MSCOCO and nocaps datasets.
    }
    \label{tab:results:caption}
\end{table}

\paragraph{Setup}
We evaluated our methods on the MSCOCO~\citep{lin-etal-2014-microsoft,karpathy-and-feifei-2015-deep} and nocaps~\citep{agrawal-etal-2019-nocaps} datasets.
We generated 256 captions per image using BLIP-2\footnote{
\texttt{Salesforce/blip2-flan-t5-xl}
}~\citep{li-etal-2023-blip-2} with epsilon sampling ($\varepsilon=0.02$) for both memory construction and decoding.
We used BERTScore\footnote{
\texttt{microsoft/deberta-v3-large}
} (\metricBertScore{})~\citep{zhang-etal-2020-bertscore,he-etal-2023-debertav3} for the utility function and evaluated captions on \metricBleu{}~\citep{papineni-etal-2002-bleu}, \metricChrf{}, and \metricBertScore{}.
In CBDT decoding, we employed DINOv2\footnote{
\texttt{facebook/dinov2-large}
}~\citep{oquab-etal-2024-dinov2} for the image similarity and \similarityMELargeInstruct{} for the caption text similarity.
We set $k=256$, $\temperature_\spaceInput=0.1$, and $\temperature_\spaceOutput=1.0$, respectively.
In memory construction, we used the training set of MSCOCO for its evaluation, and Localized Narratives (LN)~\citep{pont-tuset-etal-2020-connecting} for the nocaps evaluation\footnote{
The nocaps dataset does not contain a training set, and it was created from the Open Images~\citep{krasin-etal-2017-openimages}.
The LN~\citep{pont-tuset-etal-2020-connecting} was also created from it; thus, we used this dataset for memory construction.
}.

\paragraph{Results}
\Cref{tab:results:caption} demonstrates the results of image captioning tasks.
\decodingMbrCbdtText{} outperformed \decodingMbr{} in  \metricBleu{}, \metricChrf{}, and \metricBertScore{} on both MSCOCO and nocaps datasets, though we used \metricBertScore{} for the utility function.
These results suggest that, in addition to the choice of utility function, the domain data used for memory construction is also important.
We further discuss the limitations and additional show experimental results when constructing the memory with out-of-domain data in \cref{sec:appendix:ood}.
From the experiments, we confirmed that our methods are also effective in multimodal generation.

\section{Discussion}
\subsection{Case study: Medical translation}
\begin{table}[t]
    \centering
    \small
    \tabcolsep 3pt
    \begin{NiceTabularX}{\linewidth}{@{}lX@{}}
        \toprule
        Input & Wie wirkt \colorbox{lightgreen}{Intelence}? \\
        Reference & How does \colorbox{lightgreen}{Intelence} work? \\
        \decodingKnn{} & How does \colorbox{lightgreen}{intelligence} work? \\
        \decodingMbr{} & How does \colorbox{lightgreen}{intelligence} work? \\
        \decodingMbrCbdtText{} & How does \colorbox{lightgreen}{Intelence} work? \\
        \bottomrule
    \end{NiceTabularX}
    \caption{
    Translation examples in the medical domain.
    Both \decodingMbr{} and \decodingMbrCbdtText{} used \metricChrf{} for the utility function.
    \colorbox{lightgreen}{Highlighted spans} indicate the difference between translations.
    Other examples are shown in \cref{sec:appendix:cases}.
    }
    \label{tab:examples:domain_translation:medical}
\end{table}

We list the examples of medical translation in \cref{tab:examples:domain_translation:medical}.
\decodingMbrCbdtText{} correctly retained the medication name ``\textit{Intelence}'', but \decodingMbr{} mistranslated it to ``\textit{intelligence}''.
Accurate translation in the medical domain, like this example, is crucial for preventing serious incidents that may threaten patient safety.

We also investigated our used memory, and found that there were 37 examples that contain ``\textit{Intelence}'' on both the input and output sides in the memory, respectively.
In contrast, there were no examples in which ``Intelence'' is translated as ``intelligence''.
Thus, when ``\textit{Intelence}'' is given, the CBDT scores of hypotheses that retain ``\textit{Intelence}'' are likely to be high.
To summarize, we confirmed that \decodingMbrCbdtText{} determines the output by utilizing the information of similar examples in the memory.

\subsection{Relationship between number of examples and output quality}
\label{sec:discussion:tradeoff}

\begin{figure}[t]
    \centering
    \includegraphics[width=0.97\linewidth]{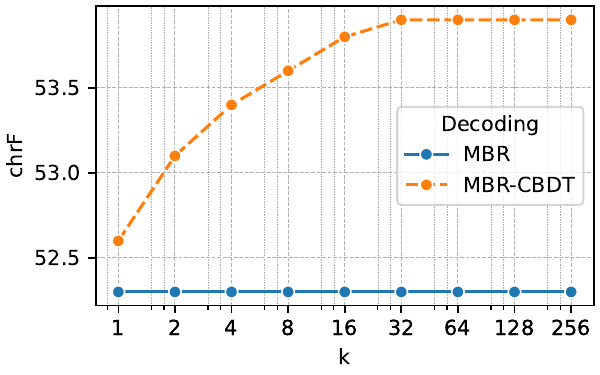}
    \caption{
    Translation quality (\metricChrf{}\%) when the number of retrieved similar examples $k$ was varied in the development set of the IT domain.
    }
    \label{fig:params:topk}
\end{figure}

\begin{figure}[t]
    \centering
    \includegraphics[width=0.97\linewidth]{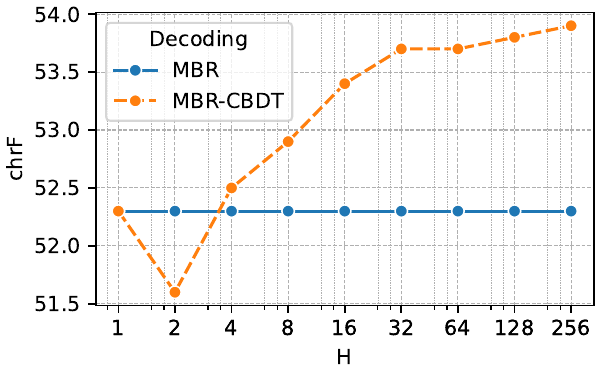}
    \caption{
    Translation quality (\metricChrf{}\%) when the number of memorized hypotheses per input $\numMemHypotheses$ was varied in the development set of the IT domain.
    }
    \label{fig:params:memory_H}
\end{figure}

We investigated the relationship between the number of examples and output quality.
\Cref{fig:params:topk} and \cref{fig:params:memory_H} show the translation quality (\metricChrf{}\%) when the number of retrieved similar examples $k$ and number of memorized hypotheses per source $\numMemHypotheses$ were varied, respectively.
As $k$ increased, memory usage increased in decoding time, but quality also improved because more examples were referenced.
$\numMemHypotheses$ also affected performance, but the quality improvement for $\numMemHypotheses \geq 16$ was less than $\numMemHypotheses < 16$.

\begin{figure}
    \centering
    \includegraphics[width=0.97\linewidth]{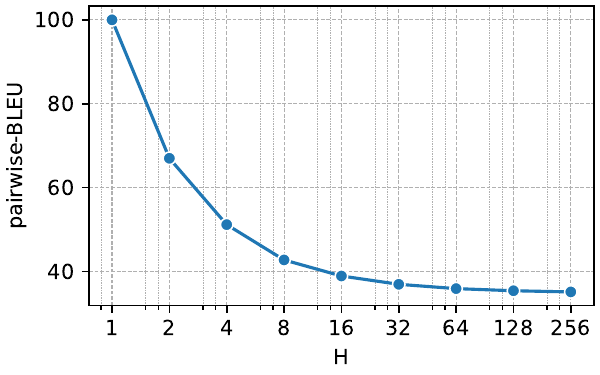}
    \caption{
    Relationship between the number of memorized hypotheses per input $H$ and the diversity of $\setHypotheses_{\textInput}$.
    A lower pairwise BLEU score means greater diversity.
    }
    \label{fig:diversity_H}
\end{figure}

We also analyzed the diversity in $\setHypotheses_{\textHypothesis}$ by evaluating the averaged pairwise BLEU~\citep{shen-etal-2019-mixture} of $\setHypotheses_{\textInput}$ on the memory in the IT domain.
\Cref{fig:diversity_H} shows the averaged scores of 1,000 random samples.
A lower score means a greater diversity of $\setHypotheses_\textInput$.
The results indicate that as $\numMemHypotheses$ increased, $\setHypotheses_\textInput$ became more diverse, but when $\numMemHypotheses \geq 16$, the pairwise BLEU converged.
These results are similar to the relationship between the number of pseudo-references and output quality in MBR decoding, i.e., a larger pseudo-reference set estimates the EU stably and converges the quality~\citep{eikema-aziz-2022-sampling,kamigaito-etal-2025-diversity,ichihara-etal-2025-theoretical}.

\subsection{Balancing MBR and CBDT decoding}
\label{sec:discussion:balance}

\begin{figure}
    \centering
    \includegraphics[width=0.97\linewidth]{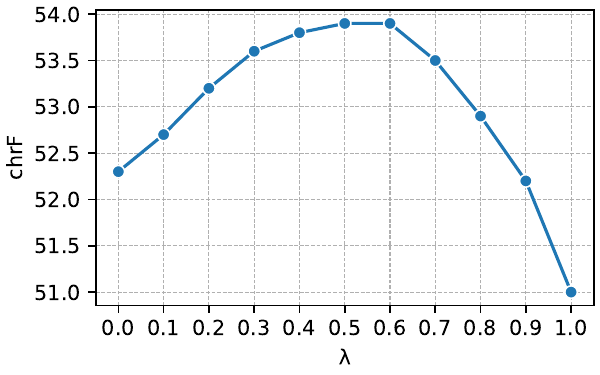}
    \caption{\metricChrf{}\% scores of $\decodingMbrCbdtText{}$ when varying the balancing weight $\lambda \in [0, 1]$ in the development set of the IT domain.}
    \label{fig:balance}
\end{figure}
We tuned the balancing weight $\lambda \in [0, 1]$.
\Cref{fig:balance} shows the \metricChrf{}\% scores of $\decodingMbrCbdtText{}$ with the various $\lambda \in \{0.0, 0.1, \ldots, 1.0\}$ in the development set of the IT domain translation.
Note that we normalized the MBR and CBDT scores by the min-max normalization, as described in \cref{eq:mbr-cbdt}.
As shown in the figure, by adjusting the weight $\lambda$, the quality can be maximized, and we selected $\lambda=0.5$ for all experiments.

\subsection{Combination of CBDT and approximated MBR decoding}
\begin{table}[t]
    \centering
    \small
    \begin{tabular}{@{}lrrrr@{}}
    \toprule
         Decoding & \metricChrf{} & \metricComet{} & \metricBleurt{} & wall-time \\
         \midrule
         \decodingMbr{} & 52.7 & 77.8 & 58.7 & 6400.3 \\
         \decodingMbrCbdtText{} & \textbf{54.6} & \underline{78.7} & \underline{59.8} & 6521.6 \\
         \styDecoding{PMbr-Cbdt} & \underline{53.2} & \textbf{79.0} & \textbf{60.2} & 1926.5 \\
         \bottomrule
    \end{tabular}
    \caption{Results of the combination of CBDT and approximated MBR decoding. ``wall-time'' indicates wall-clock time of averaged running time per test case (msec) over 5 runs.}
    \label{tab:PMBR-CBDT}
\end{table}
$\decodingMbrCbdtText{}$ takes at least as long as $\decodingMbr{}$ to execute in principle, and running time is dominated by the MBR term.
Recently, approximated methods for faster MBR decoding have been proposed~\citep{denero-etal-2009-fast,vamvas-sennrich-2024-linear,deguchi-etal-2024-centroid,cheng-vlachos-2023-faster,jinnai-ariu-2024-hyperparameter,trabelsi-etal-2024-efficient}, and we investigate the effectiveness of combining CBDT with them.
\Cref{tab:PMBR-CBDT} shows the results of the combination of CBDT decoding and probabilistic MBR decoding (\styDecoding{PMbr})~\citep{trabelsi-etal-2024-efficient}, which reduces the number of utility function calls.
In \styDecoding{PMbr}, we reduced the number of utility function calls by a factor of 64 and decomposed the pairwise score matrix with 8-dimensional two low-rank matrices.
The results indicate that \styDecoding{PMbr-Cbdt} ran 3 times faster than \decodingMbr{} and \decodingMbrCbdtText{}, yet it still outperformed \decodingMbr{} in all evaluation metrics.
Interestingly, \styDecoding{PMbr-Cbdt} outperformed \decodingMbrCbdtText{} in \metricComet{} and \metricBleurt{} with a cheaper computational cost.
This may have been owing to the mitigation of overfitting with PMBR, but further analyses remain for future work.

\subsection{Choice of similarity functions}
\begin{table}[t]
    \centering
    \small
    \begin{tabular}{@{}lrrr@{}}
    \toprule
    Similarity & \metricChrf{} & \metricComet{} & \metricBleurt{} \\
    \midrule
    \similarityMELargeInstruct{} & \textbf{54.6} & \textbf{78.7} & \textbf{59.8} \\
    \similarityLaBSE{} & \underline{54.0} & \underline{78.5} & \underline{59.1} \\
    \similarityBM{} & 51.2 & 76.5 & 57.4 \\
    \bottomrule
    \end{tabular}
    \caption{
    Comparisons of similarity functions in the IT domain translation.
    }
    \label{tab:similarity}
\end{table}

To clarify how the choice of similarity function affects the quality of output texts, we compared similarity functions and the translation quality in the IT domain.
\Cref{tab:similarity} compares similarity functions: the cosine similarity of \similarityMELargeInstruct{} and \similarityLaBSE{}~\citep{feng-etal-2022-language}, and \similarityBM{}~\citep{jones-etal-2000-probabilistic} implemented by \bms{}~\citep{lu-2024-bm25s}.
We observed that the cosine similarity of contextualized embeddings achieved higher-quality than the lexical similarity, \similarityBM{}.
Furthermore, \similarityMELargeInstruct{}, which has strong correlations with human assessment on the semantic textual similarity (STS) task~\citep{muennighoff-etal-2023-mteb}, outperformed LaBSE.
These experiments suggest that a better similarity function yields higher-quality texts.

\section{Related Work}
\paragraph{Reranking}
To enhance output quality, various reranking methods have been proposed.
One involves using different probability distributions from the generation probability.
\citet{liu-etal-2016-agreement} and \citet{imamura-sumita-2017-ensemble} used right-to-left generation models to rescore the hypotheses generated with left-to-right generation models.
Another method uses backward probabilities,
known as noisy channel decoding (NCD).
NCD first appeared in statistical machine translation~\citep{brown-etal-1990-statistical,koehn-etal-2003-statistical}, and its effectiveness has also been demonstrated in reranking of neural text generation~\citep{yu-etal-2017-neural,yee-etal-2019-simple,ng-etal-2019-facebook}.
All of the above are ``generative reranking'' based on the likelihood calculated using text generation models.
In contrast, discriminative reranking~\citep{shen-etal-2004-discriminative,lee-etal-2021-discriminative} directly distinguishes between good and bad texts and optimizes to rank the hypotheses.
The advantage of the discriminative approach is that it directly maximizes the evaluation metrics.
This means that developing metrics that accurately estimate quality will directly lead to improvements in text generation.
Quality estimation (QE)~\citep{kim-lee-2016-recurrent,kim-etal-2017-predictor,zheng-etal-2021-self} evaluates the quality of generated texts without the reference texts.
Recent QE models employ large pre-trained encoder models or large language models and achieved high correlation with human assessments~\citep{rei-etal-2022-cometkiwi,guerreiro-etal-2024-xcomet,juraska-etal-2024-metricx,li-etal-2024-llm}.

\paragraph{MBR decoding}
While most reranking methods score multiple hypotheses independently, MBR decoding~\citep{goel-and-byrne-2000-minimum,kumar-byrne-2004-minimum,eikema-aziz-2020-map,eikema-aziz-2022-sampling,freitag-etal-2022-high,fernandes-etal-2022-quality,lyu-etal-2025-unveiling} calculates the expected utility using a set of candidates.
Typically, pseudo-references sampled from the text generation model are used as a proxy for the reference texts, and their distribution affects the output quality~\citep{ohashi-etal-2024-true,kamigaito-etal-2025-diversity}.
\citet{daheim-etal-2025-uncertainty} improved the robustness of MBR decoding by using multiple generator models.
CBDT and MBR-CBDT decoding use true references in example data, not pseudo-references.

One of the major challenges of MBR decoding is that it takes quadratic time proportional to the number of candidates during inference.
To tackle this issue, efficient variants of MBR decoding have been proposed; however, they still have limitations in the evaluation metrics that can be applied~\citep{denero-etal-2009-fast,vamvas-sennrich-2024-linear,deguchi-etal-2024-centroid}, or cannot avoid on-the-fly evaluation even if the utility calling is reduced~\citep{cheng-vlachos-2023-faster,jinnai-ariu-2024-hyperparameter,trabelsi-etal-2024-efficient}.
CBDT decoding does not calculate the utility during decoding, making it faster to select the hypothesis regardless of a utility function.

\paragraph{Example-based generation}
Example-based generation is particularly useful for domain adaptation, especially in scenarios where example databases are hot-swapped.
Like other methods, CBDT decoding generates texts that reflect domain-specific information without additional training.
The idea of example-based generation originated from analogy-based machine translation~\citep{nagao-1984-framework}.
\citet{gu-etal-2018-search,zhang-etal-2018-guiding} incorporated the information of similar examples retrieved from bilingual translation memories into neural machine translation models.
Non-parametric domain adaptation using monolingual translation memory has also been proposed~\citep{cai-etal-2021-neural}.
Neural fuzzy repair~\citep{bulte-tezcan-2019-neural,xu-etal-2022-boosting,nieminen-etal-2025-incorporating} and retrieve-edit-rerank~\citep{hossain-etal-2020-simple} retrieve similar translations from translation memories.
These methods augment the input sequence with similar examples, while CBDT decoding does not.
Thus, CBDT decoding can be applied to tasks where similar examples cannot be directly concatenated with an input sequence, such as an image captioning task.
$k$NN-MT~\citep{khandelwal-etal-2021-nearest} retrieves translation examples at the token level, and interpolates the output probability based on
the distance between the current hidden representation and its nearest neighbors.
CBDT decoding mainly differs from $k$NN-MT in three key respects: the search unit, i.e., token level for $k$NN-MT and text level for CBDT decoding, the use of input side similarity, and the incorporation of utility functions.

\paragraph{Pointwise Hilbert--Schmidt independence criterion}
\citet{yokoi-etal-2018-pointwise} proposed pointwise Hilbert--Schmidt independence criterion (PHSIC), a kernel-based co-occurrence measure.
PHSIC relaxes the indicator functions on the input and output sides in pointwise mutual information (PMI) by using kernel functions.
It has been shown to be effective for selecting data from parallel corpora in translation tasks~\citep{yokoi-etal-2018-pointwise,kiyono-etal-2020-tohoku}.
Our method multiplies the input and output side similarities to soften the indicator function in \cref{eq:CBDT:naive}, and measures the value of each example in the memory.

PHSIC and CBDT decoding are mathematically similar but differ in several key respects.
Specifically, PHSIC is estimated as follows:
\begin{align}
    &\mathrm{PHSIC}(\textInput, \textOutput; \setData) \nonumber \\
    &\coloneqq \frac{1}{|\setData|}\sum_{(\tilde{\textInput}, \tilde{\textOutput}) \in \setData} \funcKernelNorm_\spaceInput(\textInput, \tilde{\textInput}) \funcKernelNorm_\spaceOutput(\textOutput, \tilde{\textOutput}),
\end{align}
where $\defFunc{\funcKernelNorm_\spaceInput}{\spaceInput \times \spaceInput}{\R}$ and $\defFunc{\funcKernelNorm_\spaceOutput}{\spaceOutput \times \spaceOutput}{\R}$ are centered kernel functions for the input and output spaces, respectively.
This closely resembles $\scoreCbdt$ in \cref{eq:CBDT:score}, but there are crucial differences.
For each input $\tilde{\textInput}$, PHSIC estimates relying solely on observed data $\tilde{\textOutput}$, whereas CBDT decoding uses multiple hypotheses $\tilde{\textHypothesis} \in \setHypotheses_{\tilde{\textInput}}$ along with their utilities $\tilde{\elmReward}$, i.e., our $\scoreCbdt$ explicitly incorporates uncertainty regarding $\tilde{\textOutput}$ through $\setHypotheses_{\tilde{\textInput}}$ and the utility $\tilde{\elmReward} = \funcUtility(\tilde{\textHypothesis}, \tilde{\textOutput})$.
In fact, when $\setHypotheses_{\tilde{\textInput}}$ contains the reference output $\tilde{\textOutput}$ and the utility function $\funcUtility$ is defined as the exact match 0--1 loss $\funcUtility(\textHypothesis, \textOutput) \coloneqq \mathbbm{1}_{\textHypothesis = \textOutput}$, the two approaches essentially become equivalent.

\section{Conclusion}
We propose CBDT decoding, which selects a high-quality hypothesis based on rewards experienced in the past to improve the quality of text generation.
The proposed method stores the rewards of hypothesis selection during memorization and uses them with similarity weights during decoding.
We further improve output quality by combining MBR and CBDT decoding.
CBDT decoding achieved better performance compared with MAP decoding, and MBR-CBDT decoding outperformed MBR decoding by up to 1.9\% in \metricChrf{} and 1.7\% in \metricComet{} in the seven domain De--En and Ja$\leftrightarrow$En translation tasks and the image captioning tasks on the MSCOCO and nocaps datasets.
We plan to investigate the effectiveness of our method in generation tasks other than text generation.

\section*{Limitations}
\paragraph{Biases from memorized data}
Our method also relies on utility functions, as does MBR decoding.
This means that the generation texts are affected by biases of utility functions.
Unlike MBR decoding, CBDT decoding is an example-based method; thus, it is susceptible to biases from memorized data.
Note that CBDT decoding does not require any additional training, so we can switch on-the-fly to use it or not.
In addition, by constructing memories for each target domain, it can hot-swap the memory according to user requests.
That is, by inserting and/or deleting examples online, such biases can be dynamically controlled.

\paragraph{Domain of memorized data}
CBDT decoding works with domain data, but when there is no domain data, it may degrade.
Specifically, we observed that the output quality became lower when constructing memory from out-of-domain data, as mentioned in \cref{sec:appendix:ood}.

\paragraph{Computational cost}
When the memorized examples are fixed, CBDT decoding works in constant time regardless of the choice of utility functions, since the utility scores are precomputed.
This efficiency comes at the cost of storage.
Specifically, the memory $\setMemory$ and intermediate representation cache of similarity functions, described in \cref{sec:prop:fastsim}, often consume large space.
This paper focused on a new paradigm for decision rules of decoding and validating its effectiveness.
Optimizing the memory and intermediate representation cache of the similarity is beyond the scope of this work, but we plan to address this limitation as future work.

\section*{Ethical Considerations}
\paragraph{Potential risks}
As mentioned in the ``Limitations'' section, CBDT decoding is susceptible to biases from memorized data.
If harmful examples are included in the memorized data and they receive high scores, harmful text is more likely to be generated.
Conversely, if we use a utility function that evaluates safe and debiased examples with high scores, CBDT decoding may suppress the output of harmful cases.
Therefore, to use CBDT decoding safely, we should develop non-toxic datasets used for the memories and utility functions that evaluate them with high scores.

\paragraph{Use of benchmark datasets}
In our experiments, we only used publicly available benchmark datasets as described in \cref{sec:exp}.
All datasets we used were created on domains that do not contain personal information or offensive content.
They are released under the licenses described in \cref{sec:licenses}, and we complied with them.


\bibliography{anthology-1,anthology-2,custom}

\appendix

\section{Hyperparameters}
\label{sec:tuning}

\begin{table}[H]
    \centering
    \small
    \begin{tabular}{@{}llrrr@{}}
    \toprule
        & & \multicolumn{3}{@{}c@{}}{$\temperature_\spaceInput$} \\
        \cmidrule(l){3-5}
        Development set & $\temperature_\spaceOutput$ & 0.01 & 0.1 & 1.0 \\
        \midrule
        IT De--En & 0.01 & \textbf{53.93} & 53.76 & 53.72 \\
                  & 0.1  & 53.69 & 53.62 & 53.60 \\
                  & 1.0  & 53.70 & 53.60 & 53.57 \\
        \midrule
        ASPEC Ja--En & 0.01 & 48.03 & \textbf{48.13} & 48.09 \\
                     & 0.1  & 47.96 & 48.02 & 47.97 \\
                     & 1.0  & 47.94 & 47.95 & 47.93 \\
        \bottomrule
    \end{tabular}
    \caption{$\metricChrf{}$ scores when temperatures varied on the development set in the De--En IT domain and Ja--En ASPEC translation tasks.}
    \label{tab:tuning:translation}
\end{table}

\begin{table}[H]
    \centering
    \small
    \begin{tabular}{@{}llrrr@{}}
    \toprule
        & & \multicolumn{3}{@{}c@{}}{$\temperature_\spaceInput$} \\
        \cmidrule(l){3-5}
        Development set & $\temperature_\spaceOutput$ & 0.01 & 0.1 & 1.0 \\
        \midrule
        MSCOCO & 0.01 & 67.93 & 68.54 & 67.93 \\
               & 0.1  & 68.38 & 68.49 & 68.31 \\
               & 1.0  & 68.25 & \textbf{68.55} & 68.53 \\
        \bottomrule
    \end{tabular}
    \caption{$\metricBertScore{}$ scores when temperatures varied on the development set in the MSCOCO image captioning task.}
    \label{tab:tuning:caption}
\end{table}

CBDT decoding has four hyperparameters: number of memorized hypotheses per input $H$, number of neighboring examples $k$, and temperatures of similarities in the input side $\temperature_\spaceInput$ and output side $\temperature_\spaceOutput$.
As described in \cref{sec:discussion:tradeoff}, $H$ and $k$ are trade-off parameters between memory usage and quality.
The other two, the temperatures of similarities $\temperature_\spaceInput$ and $\temperature_\spaceOutput$, need to be tuned.

In our experiments, we tuned them from $\{0.01, 0.1, 1.0\}$ on the development set.
Intuitively, lower temperatures emphasize the similarity scores.
In the De--En domain translation tasks, we tuned them on the development set of the IT domain so that \metricChrf{} is maximized.
Likewise, in the Ja$\leftrightarrow$En domain translation tasks, we maximized \metricChrf{} on the development set of the ASPEC Ja$\rightarrow$En.
In the image captioning tasks, we tuned them on the development set of the MSCOCO dataset by maximizing BERTScore (\metricBertScore{}).
The results of development sets when the temperatures varied in the translation and image captioning tasks are shown in \cref{tab:tuning:translation} and \cref{tab:tuning:caption}, respectively.

\section{Dataset Statistics}

\begin{table}[H]
    \centering
    \small
    \begin{tabular}{@{}lrrr@{}}
        \toprule
        Dataset & Train & Dev & Test \\
        \midrule
        \multicolumn{4}{@{}l@{}}{\textit{De--En domain translation tasks}} \\
        ~~~IT & 222,927 & 2,000 & 2,000 \\
        ~~~Koran & 17,982 & 2,000 & 2,000 \\
        ~~~Law & 467,309 & 2,000 & 2,000 \\
        ~~~Medical & 248,099 & 2,000 & 2,000 \\
        ~~~Subtitles & 500,000 & 2,000 & 2,000 \\
        \midrule
        \multicolumn{4}{@{}l@{}}{\textit{Ja$\leftrightarrow$En domain translation tasks}} \\
        ~~~ASPEC & 1,000,000 & 1,790 & 1,812 \\
        ~~~KFTT & 440,288 & 1,166 & 1,160 \\
        \midrule
        \multicolumn{4}{@{}l@{}}{\textit{Image captioning tasks}} \\
        ~~~MSCOCO & 113,287 & 5,000 & 5,000 \\
        ~~~nocaps & \tablefootnote{We used Localized Narratives~\citep{pont-tuset-etal-2020-connecting} as described in \cref{sec:exp:caption}.}504,413 & 4,500 & \tablefootnote{
        We did not use the test set but used the development set for the nocaps evaluation because the reference captions of the test set are not publicly available.
        Thus, we tuned the hyperparameters for both MSCOCO and nocaps on the development set of MSCOCO.
        }(10,600) \\
        \bottomrule
    \end{tabular}
    \caption{
    Number of examples for each dataset.
    Note that we used training sets for memory construction.
    }
    \label{tab:datastats}
\end{table}

\Cref{tab:datastats} shows the number of examples for each dataset.

\section{Further Analyses}
\subsection{Relationship between memory domain and output quality}
\label{sec:appendix:ood}

\begin{table}[H]
    \centering
    \small
    \tabcolsep 3pt
    \begin{tabular}{@{}lrrr@{}}
    \toprule
        Memorized dataset & \metricBleu{} & \metricChrf{} & \metricBertScore{} \\
        \midrule
        LN (target domain) & 26.8 & 41.6 & 66.3 \\
        MSCOCO (non-target domain) & 25.7 & 41.7 & 66.0 \\
        \bottomrule
    \end{tabular}
    \caption{Results of MBR-CBDT decoding with different domain memories in the nocaps image captioning task.}
    \label{tab:ood}
\end{table}

We investigated the effectiveness of using target domain data for memory construction.
As mentioned in \cref{sec:exp:caption}, we constructed the memory using the Localized Narratives (LN) dataset~\citep{pont-tuset-etal-2020-connecting} for the image captioning task on the nocaps dataset.
We compared this with constructing the memory from the training set of the MSCOCO dataset, which differs from the target domain.
\Cref{tab:ood} shows the results.
When using non-target domain data for memory construction, MBR-CBDT degraded in \metricBleu{} and \metricBertScore{}.
These results indicate that MBR-CBDT decoding is effective when the memory contains similar examples for the current input.

\subsection{Additional case studies}
\label{sec:appendix:cases}
\begin{table}[H]
    \centering
    \small
    \tabcolsep 3pt
    \begin{NiceTabularX}{\linewidth}{@{}lX@{}}
        \toprule
        Input & Insulin Human Winthrop Comb 50 ist eine Fl\"ussigkeit (Suspension) \colorbox{lightgreen}{zum Spritzen unter die} \colorbox{lightgreen}{Haut}. \\
        Reference & Insulin Human Winthrop Comb 50 is a fluid (suspension) \colorbox{lightgreen}{for injection under the skin}. \\
        \decodingKnn{} & Insulin Human Winthrop Comb 50 is a liquid (suspension) \colorbox{lightgreen}{for injection under the skin}. \\
        \decodingMbr{} & Insulin Human Winthrop Comb 50 is a liquid (suspension) \colorbox{lightgreen}{to be sprayed under the skin}. \\
        \decodingMbrCbdtText{} & Insulin Human Winthrop Comb 50 is a liquid (suspension) \colorbox{lightgreen}{for injection under the skin}. \\
        \bottomrule
    \end{NiceTabularX}
    \caption{
    Translation examples in the medical domain.
    Both \decodingMbr{} and \decodingMbrCbdtText{} used \metricChrf{} for the utility function.
    \colorbox{lightgreen}{Highlighted spans} indicate the difference between translations.
    }
    \label{tab:examples:domain_translation:medical-511}
\end{table}

We also present examples of medical translation in \cref{tab:examples:domain_translation:medical-511}.
\decodingMbrCbdtText{} correctly translated ``\textit{zum Spritzen unter die}'' to ``\textit{for injection under the skin}``, while \decodingMbr{} translated it to ``\textit{to be sprayed under the skin}''.
$k$NN-MT generated a mistranslation in \cref{tab:examples:domain_translation:medical}, but it succeeded in this case.
In this example, there were 11 instances in the memory where the input text contained ``\textit{zum Spritzen unter die}'' and the corresponding reference contained ``\textit{for injection under the skin}''.
Similar to \cref{tab:examples:domain_translation:medical}, there were no memory instances where the reference text contained ``\textit{to be sprayed under the skin}'' when the input text contained ``\textit{zum Spritzen unter die}''.

\section{Licenses}
\label{sec:licenses}
\begin{table*}[t]
    \centering
    \small
    \begin{tabular}{@{}lll@{}}
        \toprule
        Model & License & Reference \\
        \midrule
        \texttt{facebook/m2m100\_418M} & MIT & \citet{fan-etal-2021-beyond} \\
        \texttt{intfloat/multilingual-e5-large-instruct} & MIT & \citet{wang-etal-2024-multilingual} \\
        \texttt{Unbabel/wmt22-comet-da} & Apache-2.0 & \citet{rei-etal-2022-comet} \\
        \texttt{Unbabel/wmt22-cometkiwi-da} & CC BY-NC-SA 4.0 & \citet{rei-etal-2022-cometkiwi} \\
        \texttt{Salesforce/blip2-flan-t5-xl} & MIT & \citet{li-etal-2023-blip-2} \\
        \texttt{facebook/dinov2-large} & Apache-2.0 & \citet{oquab-etal-2024-dinov2} \\
        \texttt{microsoft/deberta-v3-large} & MIT & \citet{he-etal-2023-debertav3} \\
        \texttt{sentence-transformers/LaBSE} & Apache-2.0 & \citet{feng-etal-2022-language} \\
        \bottomrule
    \end{tabular}
    \caption{Licenses of models we used.}
    \label{tab:licenses:models}
\end{table*}

\paragraph{Datasets}
The five domain De--En translation datasets can be used for research purposes as described in the original paper~\citep{koehn-knowles-2017-six,aharoni-goldberg-2020-unsupervised}.
ASPEC~\citep{nakazawa-etal-2016-aspec} can be used for research purposes as described in \url{https://jipsti.jst.go.jp/aspec/}.
KFTT~\citep{neubig-2011-kftt} is licensed by Creative Commons Attribution-Share-Alike License 3.0.
In the MSCOCO dataset~\citep{lin-etal-2014-microsoft,karpathy-and-feifei-2015-deep}, the annotations belong to the COCO Consortium and are licensed under a Creative Commons Attribution 4.0 license.
The COCO Consortium does not own the copyright of the images, and use of the images must abide by the Flickr Terms of Use.
The nocaps dataset~\citep{agrawal-etal-2019-nocaps} is released under a CC-BY-2.0 License, and the Localized Narratives dataset~\citep{pont-tuset-etal-2020-connecting} is released under a CC-BY-4.0 License.

\paragraph{Models}

The licenses of models we used are listed in \cref{tab:licenses:models}.

\paragraph{Tools}
We used the following MIT-licensed tools.
\begin{itemize}
    \item \mbrs{}~\citep{deguchi-etal-2024-mbrs}
    \item \knntransformers{}~\citep{alon-etal-2022-neuro}
    \item \faiss{}~\citep{johnson-etal-2019-billion,douze-etal-2024-faiss}
    \item \bms{}~\citep{lu-2024-bm25s}
\end{itemize}

\end{document}